%% file: jvta-coling.tex
\documentclass[11pt]{article}
\usepackage{coling2018}
\usepackage{times}
\usepackage{url}
\usepackage{latexsym}
\usepackage{xcolor}
\usepackage{soul}
\usepackage[utf8]{inputenc}
\usepackage{graphicx}
\usepackage{bm}
\usepackage{amsmath}
\usepackage{multirow}
\usepackage{booktabs}
\usepackage{comment}
\usepackage{titlesec}
\usepackage{wrapfig}
\usepackage{enumitem}

\title{JTAV: Jointly Learning Social Media Content Representation by Fusing\\
       Textual, Acoustic, and Visual Features}
\blfootnote{
	 \hspace{-0.65cm}  % space normally used by the marker
	 This work is licensed under a Creative Commons 
	 Attribution 4.0 International License.
	 License details:
	 \url{http://creativecommons.org/licenses/by/4.0/}
}
\author{Hongru Liang\textsuperscript{1},
	Haozheng Wang\textsuperscript{1},
    Jun Wang\textsuperscript{2},
	Shaodi You\textsuperscript{3},\\
    \textbf{Zhe Sun}\textsuperscript{4},
	\textbf{Jin-Mao Wei}\textsuperscript{1},
	\textbf{Zhenglu Yang}\textsuperscript{1}\thanks{~~Corresponding author} \\ 
	\textsuperscript{1} CCCE, Nankai University, Tianjin, China \\
    \textsuperscript{2} College of Mathematics and Statistics Science, Ludong University, China\\	
	\textsuperscript{3} Data61-CSIRO, Australian National University, Canberra, Australia\\	
    \textsuperscript{4} RIKEN Head Office for Information Systems and Cybersecurity\\ 
                        Computational Engineering Applications Unit, Japan\\	
	{\{lianghr, hzwang, junwang\}@mail.nankai.edu.cn}, {shaodi.you@data61.csiro.au},\\
zhe.sun.vk@riken.jp,    {\{weijm, yangzl\}@nankai.edu.cn}
}

\begin{document}
\maketitle
\begin{abstract}
Learning social media content is the basis of many real-world applications, including information retrieval and recommendation systems, among others. In contrast with previous works that focus mainly on single modal or bi-modal learning, we propose to learn social media content by fusing \underline{j}ointly \underline{t}extual, \underline{a}coustic, and \underline{v}isual information (JTAV). Effective strategies are proposed to extract fine-grained features of each modality, that is, attBiGRU and DCRNN. We also introduce cross-modal fusion and attentive pooling techniques to integrate multi-modal information comprehensively. Extensive experimental evaluation conducted on real-world datasets demonstrates our proposed model outperforms the state-of-the-art approaches by a large margin.
\end{abstract}

\input{body}
% \clearpage
\bibliographystyle{acl}
\bibliography{coling2018}
\end{document}

%% file: body.tex
\section{Introduction}
\label{sec:intro}
\begin{figure}[h]
	\centering
	\includegraphics[width=0.6\textwidth]{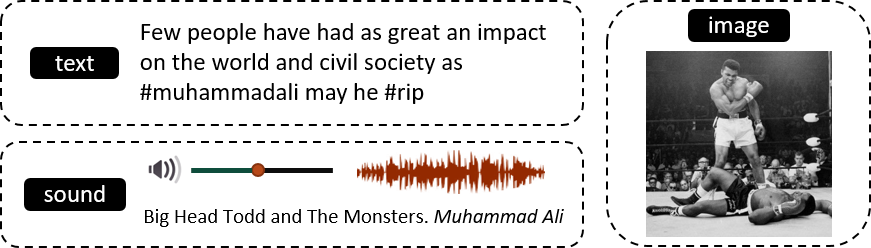}
	\caption{An example of multi-modal social media content, organized by a textual passage, a clip of the song ``Muhammad Ali'', and a picture of Muhammad Ali's famous victory.}
	\label{fig:example}
\end{figure}

The popularity of the social media (e.g., Twitter, Weibo) over the last two decades has led to increasing demands for learning the content of social media, which may be beneficial to many real-world applications, such as sentiment analysis~\cite{wang2015unsupervised}, information retrieval~\cite{li2017image2song}, and recommendation systems~\cite{wu2017mobile}. However, the task is non-trivial and challenging, because social media content is commonly multi-modal and involves several types of data, including text, audio, and image (an example is illustrated in Fig.~\ref{fig:example}). Each representation of the individual modality encodes specific knowledge and complementary, which can be explored to facilitate understanding of the entire meaning of the content. Therefore, gaining comprehensive knowledge from learning arising out of social media requires deliberate exploration of single-modal information and joint learning of the intrinsic correlation among various-modalities.\par
%\begin{wrapfigure}{r}{10cm}
%	\centering
%	\includegraphics[width=0.6\textwidth]{example.png}
%	\caption{An example of multi-modal social media content, organized by a textual passage, a clip of the song ``Muhammad Ali'', and a picture of Muhammad Ali's famous victory.}
%    \label{fig:example}
%\end{wrapfigure}
Previous works on learning social media content have focused mainly on single modality, e.g., textual information~\cite{wu2017mobile} and image information~\cite{garimella2016social}. To our best knowledge, no studies have focused on mere acoustic information analysis for social media. These previous works cope with data in the early stage of social media, because users are limited to posting single-modal content. In contrast, today's social media content usually includes textual, acoustic, and visual ingredients simultaneously, and requires multi-modal learning strategies to integrate rich information from all modalities.\par

Many researchers have studied the bi-modal utilization to address the issue of incomplete information exploration. This utilization can be classified into two kinds of approaches. The first is bi-modal retrieval, which utilizes one modality to find similar content in the other modality, such as finding the most relevant texts to a given image~\cite{li2017image2song}. This kind of approach is similar to machine translation tasks to some extent, for instance, the input is fine-grained features of one modality, the objective of which is to minimize the distance between the output vectors with fine-grained features of the target modality. The second is bi-modal unification~\cite{park2016image}, which seeks to integrate individual information of the two modalities together. The process of this kind of approach can be divided into two stages: (a) feature extraction, in which the two modalities are encoded into their corresponding vectors or matrices via embedding approaches; and (b) bi-modal feature aggregation, in which the extracted features are mapped into a shared latent space through multilayer perceptrons or directly into the target spaces of tasks through classifiers~\cite{chen2017visual}. \par
Learning representations of multi-modal data, which contains features of more than two modalities, is a more challenging task. The main reason is that the fine-grained features of the modalities can be difficult to obtain because of the lack of annotated labels for every modality. Previous studies lack the appropriate strategies to aggregate various kinds of information effectively. It's easy to imagine that bi-modal aggregation models might be extended to multi-modal learning with linear operations, e.g., the concatenation operation. However, this technique may generate inappropriate integral representation of the multiple features and lacks further exploration in the previous work. \par

We address the abovementioned issues by introducing effective strategies to represent the individual features for the three modalities. A general unified framework is proposed to integrate multi-modalities seamlessly. Specifically, we begin by encoding single modal parts into dense vectors. For textual content, we design an attention-based network~(i.e., attBiGRU), to incorporate various textual information. For acoustic content, we introduce an effective DRCNN approach to embed temporal audio clips locally and globally. For visual content, we fine-tune a state-of-the-art general framework, DenseNet~\cite{huang2017densely}. We propose a novel fusion framework, which involves the cross-modal fusion and the attentive pooling strategies, to aggregate various modal information. Extensive experimental evaluations conducted on real-world datasets demonstrate that our proposed model outperforms state-of-the-art approaches by a large margin. The contributions of this paper are presented as follows:

%The matrix multiplications are built on dense vectors of all modalities pairwise, and we obtain a suite of cross-modal matrices. The attentive pooling strategy is introduced to reconstruct the individual modal feature with cross-modal wise. The integration of reconstructed vectors is used as the final representation, which encodes the inner information of single modal with the outer relevant information in cross-modal sense.\par

%We name the whole multi-modal learning framework as JVTA; ``J'' for jointly learning, ``V'' for visual part, ``T'' for textual part, and ``A'' for acoustic part. 

%We highlight our contributions in four aspects:
\begin{itemize}
	
	\item We introduce effective strategies to extract representative features of the three modalities, including the following. (1) For textual information, we design an attention based network, named as attBiGRU, to integrate various textual parts. The independent textual parts are considered as two separate roles, namely, the protagonist and the supporting players. The supporting players are encoded into an attention weight vector on the protagonist. (2) For acoustic information, we propose the DCRNN strategy based on the densely connected convolutional and recurrent networks. The densely connected convolutional networks are used to learn the acoustic features both locally and globally. The recurrent networks are designed to learn the temporal information. (3) For visual information, we introduce state-of-the-art strategies from~\cite{simonyan2014very,he2016deep,huang2017densely}.
    
   \item We propose a general multi-modal feature aggregation framework, JTAV, to learn information jointly. The framework considers the textual, acoustic, and visual contents to generate a unified representation of social media content with the help of the optimized feature fusion network, CMF-AP, which can learn inner and outer cross-modal information simultaneously.
%The learned representation incorporates each modal information comprehensively, and generates state-of-the-art results on experiments.
   
    \item We conduct comprehensive experiments on two kinds of tasks, sentiment analysis and information retrieval, to evaluate the effectiveness of the proposed JTAV framework. The experiments demonstrate that JTAV outperforms state-of-the-art approaches remarkably, that is, about 2\% higher than the state-of-the-art approaches on all metrics in the sentiment analysis experiment and more than ten times better than the baseline on main metrics in music information retrieval experiment.
    
\end{itemize}

%Text-aware representation of posts has been intensively studied
\section{Related Work}
Over the past decades, social media learning has gained considerable attention in related research literature. Social media content is utilized as raw material for recommendation~\cite{wang2013social}, information retrieval~\cite{agichtein2008finding}, and so forth.\par
The dominance of the single-modal content in social media in the early stages, however, has caused most previous works to position themselves in single-modal exploration, e.g., natural language processing and image processing. \cite{han2013lexical} focused on the short text messages in social media, and normalized lexical variants to their canonical forms. In~\cite{hutto2014vader}, combined lexical features of social media text were proposed for expressing and emphasizing sentiment intensity. An example of image processing was ~\cite{garimella2016social}, social media images, particularly geo-tagged images, were studied in predicting county-level health statistics.\par

In recent years, users have come to prefer presenting more attractive ingredients, such as image and audio in addition to natural words. Therefore, literature on social media analysis has been shifting its focus from single-modal to multi-modal learning. For example, ~\cite{wang2015unsupervised} assume that visual and textual content have a shared sentiment label space, and the optimization problem can be converted to the minimization between the derivatives of both modalities. In~\cite{li2017image2song}, the images and lyrics are mapped into a shared tag space. The lyrics, which have the smallest mean squared error with a given image, emerged as the matching object. \cite{chen2017visual} present a new end-to-end framework for visual and textual sentiment analysis; they began with the co-appearing image and text pairs that provide semantic information for each other, and use the concatenation of the vectors generated from the original image and text, in the shared sentiment space. However, the drawback of these works is that they utilized only partial information.\par

An intuitive idea to gain a comprehensive learning of the entire meaning of social media is to integrate more modalities effectively. The reason is that each representation of the individual modality encodes specific knowledge and is complementary, an aspect that can be explored to facilitate understanding on the entire meaning of the content. However, this task could be extremely challenging because we need to explore single-modal information deliberately and jointly learn the intrinsic correlation among various modalities. This work is the first study that focuses on the seamless integration of the three modalities~(i.e., text, audio, and image), into a general framework to jointly learn the social media content.\par

\section{Model Description}
\begin{figure}[t]
	\centering
	\includegraphics[width=0.66\textwidth]{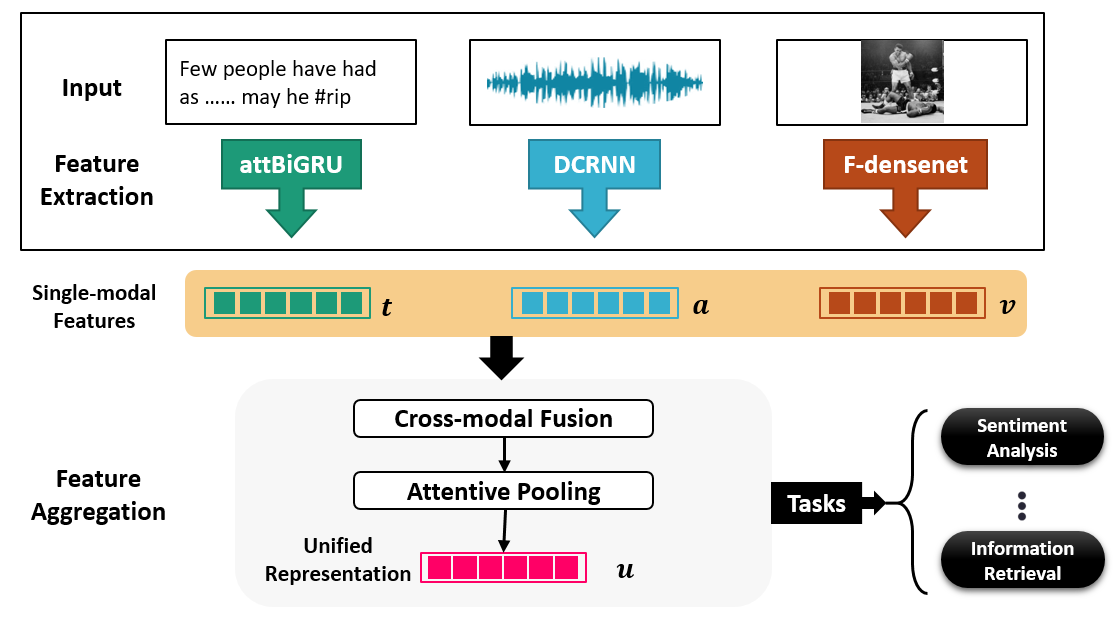}
	\caption{The architecture of the proposed JTAV framework}
	\label{fig:model}
\end{figure}

The architecture of our proposed framework is illustrated in Fig.~\ref{fig:model}. It has two main modules, the feature extraction and the feature aggregation. Let $\bm t$, $\bm a$, and $\bm v$ denote the textual, acoustic, and visual features, which are encoded from the feature extraction module, respectively. Let $\bm u$ denote the target unified representation of the multi-modal content. The feature aggregation module is designed to generate $\bm u$ from $\bm t$, $\bm a$, and $\bm v$ effectively and comprehensively. The example in Fig.~\ref{fig:example} is utilized to ground our discussion and to illustrate the intuitive idea of this work.
\subsection{Feature Extraction}
Fine-grained features, which are related to the qualities of the input of the aggregation processing directly, are the foundation of multi-modal representation learning. The proposed strategies for generating appropriate features for text, audio, and image will be presented in the following subsections. \par
\subsubsection{Text modeling}
\label{sec:text}
%\begin{wrapfigure}{l}{8cm}
%    \centering
%	\includegraphics[width=8cm]{text.png}
%	\caption{The attBiGRU strategy for modeling textual content }	
%	\label{fig:text}
%\end{wrapfigure}
\begin{figure}[t]
	\centering
	\includegraphics[width=0.6\textwidth]{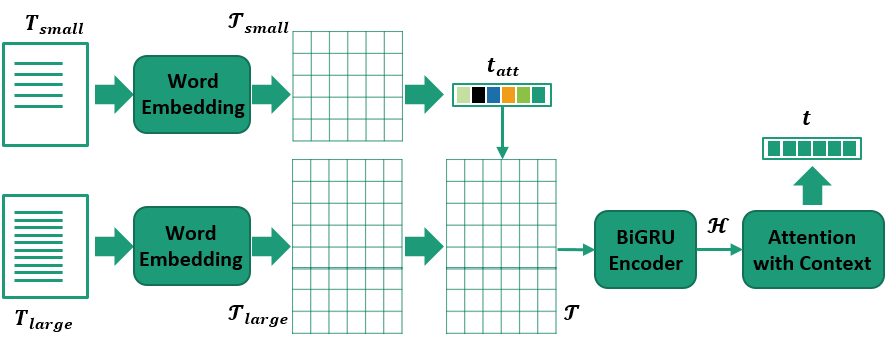}
	\caption{The attBiGRU strategy for modeling textual content }	
	\label{fig:text}
\end{figure}
The purpose of text modeling is to encode raw textual parts into a latent representation $\bm t$. Inspired by \cite{lei2017swim,lei2018sequicity,lei2018linguistic}, we design a bidirectional gated recurrent neural network (BiGRU) based attentive network~(i.e., attBiGRU), to extract sufficient textual features carrying the long-range dependencies, as illustrated in Fig.~\ref{fig:text}.\par
Generally, social media content contains more than one textual part, such as the lyrics and reviews of a song. It is reasonable to assume that a protagonist and supplying players exist among these textual parts. Intuitively and for sake of simplicity, we can regard the part with the maximum number of words as the protagonist and the other parts as the supplying players. Let $\bm {T_{large}}$ denote the protagonist and $\bm {T_{small}}$ denote the supplying roles. Inspired by~\cite{li2017image2song} to reduce the gap between image and text, we use a pre-trained word embedding model, e.g., FastText~\cite{bojanowski2016enriching}, to encode $\bm {T_{small}}$ into a word matrix~(i.e., $\bm{\mathcal{T}_{small}}$). $\bm{\mathcal{T}_{small}}$ is computed as $\bm{w_{small}} W_e$, where $\bm{w_{small}}$ is the words in $\bm {T_{small}}$ and $W_e$ is a pre-trained embedding matrix. $\bm {t_{att}}$ is an attention vector generated from $\bm{\mathcal{T}_{small}}$ with an average pooling layer, and works on every word vector of $\bm{\mathcal{T}_{large}}$ through the element-wise product. We obtain an attented word matrix, denoted as $\bm{\mathcal{T}}$. \par
The BiGRU encoder is utilized to encode every word vector in the sequence into a latent representation and receive the aid of the other textual parts. Subsequently, we feed the attended word matrix $\bm {\mathcal{T}}$ into a BiGRU network. The output of the BiGRU encoder is a $2M\times N$ matrix, where $M$ is the number of hidden units of a layer, and $N$ is the number of words in $\bm{\mathcal{T}_{large}}$. We denote the output matrix as $\bm {\mathcal{H}}$.\par
The attention with context part is a simplified version of~\cite{yang2016hierarchical} without the sentence attention part. It is introduced to balance the importance among all words and is defined as follows:
\begin{center}
\begin{equation}
\begin{split}
\label{equ:text}
\bm{\hat{h}_i}&= tanh(\bm {W_u}\bm{h_i}+\bm{b_u}),\\
\bm{\alpha_i}&= \frac{exp\left( \bm{\hat{h}{_i}^{\top } \hat{h}_{c}}\right) }{\sum_{j=0}^{N-1}{exp\left( \bm {\hat{h}{_j}^\top \hat{h}_{c}} \right) } },\\
\bm{t}&= \sum_{j=0}^{N-1}{\bm{\alpha{_j} h{_j}}},
\end{split}
\end{equation}
\end{center}
where $\bm {h_i}(0\le i \le N - 1)$ is the $i^{th}$ column of $\bm{\mathcal{H}}$, $\bm {\hat{h}_i} $ is a latent vector computed from $\bm h_i$ using the tanh function, $\bm {\alpha_i}$ is the attention weight scalar computed from $\bm {\hat{h}_i}$ and a random initialized context vector $\bm{\hat{h}_{c}}$, $\bm {W_u}$ and $\bm {b_u}$ are parameters learned from training. The final textual representation $\bm t$ is a weighted sum of $\bm {h_j}(0\le j \le N - 1)$ on $\bm {\alpha_j}(0\le j \le N - 1)$.\par
\subsubsection{Audio modeling}
%\begin{wrapfigure}{r}{12cm}
%	\centering
%	\includegraphics[width=8cm]{audio.png}
%	\caption{The DRCNN strategy for modeling acoustic content }
%	\label{fig:audio}
%\end{wrapfigure}
\begin{figure}[t]
	\centering
	\includegraphics[width=0.7\textwidth]{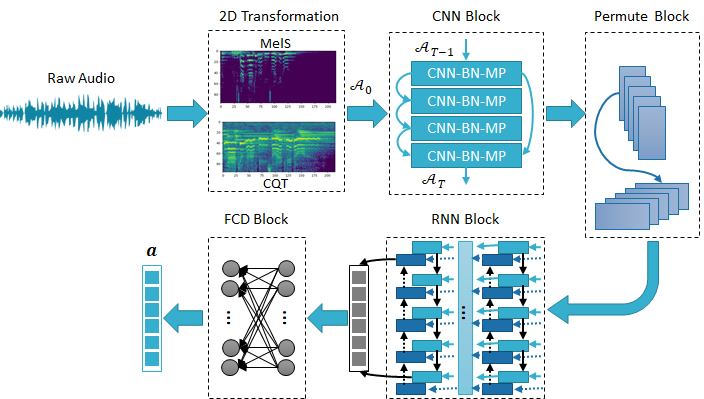}
	\caption{The DRCNN strategy for modeling acoustic content }
	\label{fig:audio}
\end{figure}
In this paper, we focus on polyphonic audio, such as music clips. To capture useful acoustic features, we design a densely connected convolutional recurrent neural network called DCRNN, which takes raw audio as input, and generates a distributed representation $\bm a$ to effectively represent the acoustic information. DCRNN consists of five phases, as illustrated in Fig.~\ref{fig:audio}.\par
The first phase is named as 2D transformation, which transfers raw audio into two-dimensional vectors. Following the majorities of the deep learning approaches in music information retrieval~\cite{choi2017tutorial}, we use two dimensional spectrograms instead of the discrete audio signals. We choose two kinds of spectrograms~(i.e., mel-spectrogram~\cite{boashash1996time} and constant-Q transform~\cite{brown1992efficient}). A mel-spectrogram~(MelS) is a visual representation of the spectrum of frequencies of audio and is optimized for human auditory perception. A constant-Q transform~(CQT) is computed with the logarithmic-scale of the central frequencies. The CQT is well suited with the frequency distribution of music pitch. MelS and CQT have been used frequently in various acoustic tasks~\cite{zhu2017fusing,aren2017towards}, and are utilized as coarse-grained features in the current paper. We use $\bm{\mathcal{A}_0}$ to indicate the 2D spectrograms.\par
The second phase is a convolutional neural network (CNN) based block~(called CNN block). $\bm{\mathcal{A}_0}$ is used as the input of the block. According to~\cite{choi2017transfer}, early layers of CNN have the ability to capture local information such as pitch and harmony, whereas deeper layers can capture global information such as melody. We introduce the densely connected CNN, which has been proved powerful at learning both local and global features of images~\cite{huang2017densely}. The CNN block is composed of several densely connected ``CNN-BN-MP'' sub-blocks, where ``BN'' represents batch normalization layers, and ``MP'' indicates max pooling layers. Suppose there are $N$  densely connected ``CNN-BN-MP'' sub-blocks in the CNN block, each can be defined as:\par
\begin{center}
\begin{equation}
\begin{split}
\bm{\mathcal{A}_h^1}&=CNN-BN-MP(\bm{\mathcal{A}_{T-1}}),\\
\bm{\mathcal{A}_h^2}&=CNN-BN-MP(\bm{\mathcal{A}_h^1}),\\
\bm{\mathcal{A}_h^3}&=CNN-BN-MP(\bm{\mathcal{A}_h^2}),\\
\bm{\mathcal{A}_{T}}&=CNN-BN-MP(\bm{\mathcal{A}_h^1}\oplus \bm{\mathcal{A}_h^3}),
\end{split}
\end{equation}
\end{center}
where $\bm T\in (1,N)$, $\bm{\mathcal{A}_{T-1}}$ represents the output of the ${ ( \bm T-1)}^{th}$ densely connected ``CNN-BN-MP'' sub-block, $\bm{\mathcal{A}_h^*}$ indicate the latent variables learned by the CNN block, $\oplus$ denotes the concatenation operation, and $\bm{\mathcal{A}_{T}}$ is the output of the $\bm T^{th}$ sub-block. \par
The permute block is adopted from the technique introduced in ~\cite{panwar2017deep}. The output of the CNN block is permuted to time major form, and fed into the following recurrent neural network.\par
The recurrent neural network~(RNN) block is built based on bidirectional RNN. The stacked RNN layers are used to learn the long-short term temporal context information. We take the concatenation of the last hidden state (in vector format) in the forward direction and the first hidden state in the backward direction as the output of this phase.\par
Several fully connected dense layers constitutes the fully connected dense~(i.e., FCD) block. We use the output of the last dense layer as the acoustic representation $\bm{a}$. Note that all the parameters presented in this section, e.g., number of layers, are optimally tuned and will be clarified in the experiment section.\par
\subsubsection{Image modeling}
Extracting  rewarding features from images is a well explored problem; hence, following the tradition of other image-related tasks~\cite{li2017image2song,chen2017visual}, we use pre-trained approaches on external visual tasks. \par
VGG~\cite{simonyan2014very} is a deep convolutional network with nineteen weight layers, ResNet~\cite{he2016deep} is a residual learning framework with 101 weight layers, and DenseNet is a deep densely connected convolutional network with 121 weight layer. All these models are pre-trained on the ImageNet dataset~\footnote{\url{http://www.image-net.org}.}, and are fine-tuned on the experimental datasets by modifying the final densely-connected layers. The last dense layer is used as the visual representation $\bm v$.

\subsection{Feature Aggregation}
In this section, we propose a feature aggregation network, which consists of the cross-modal fusion~(CMF) and the attentive pooling~(AP) strategies to generate a unified representation of multi-modal information, as shown in the bottom part of Fig.~\ref{fig:model}. We call the feature aggregation approach as CMF-AP.\par

The first phase is cross-modal fusion. We introduce the concepts of the inner and outer information. The inner information refers to information remaining in single modality, like the visual information of an image or the acoustic information of a song clip. The outer mutual information cross modalities refers to the relevant information between two modalities. In addition to the inner information generated from the feature extraction module, we compute the matrix product ($\times$) pairwise on all modal vectors to calculate the cross-modal aware matrices. The cross-modal aware matrices contain outer mutual information cross modalities and the inner information. \par

However, handling the cross-modal aware matrices directly is not feasible, and thus, we reconstruct $\bm t$, $\bm a$, and $\bm v$ from the cross-modal aware matrices. The attentive pooling strategy is adapted from Eq.~\ref{equ:text} to conduct row pooling and column pooling.\par

After the above procedure, two reconstructed $\bm t$, two reconstructed $\bm a$, and two reconstructed $\bm v$ are obtained. For instance, from the text-image matrix, we can obtain a text-enhanced visual vector and an image-enhanced textual vector. The reconstructed vectors carry both inner and outer information. Densely connected layers are deployed for dimensionality reduction purpose. The final unified representation $\bm u$ is the concatenation of the reconstructed vectors. \par
The feature aggregation module is formulated as follows:
\begin{center}
  \begin{equation}
    \begin{split}
      \begin{gathered}
        \label{equ:cmf}
        {\bm {C_{mn}}}=\bm m \times \bm n^\top ;\\
        \begin{aligned}
          &\left\{
          \begin{aligned}
          & {\bm {C_m}}=tanh(\bm {W_m}\bm {C_{mn}}+\bm {b_m}),\cr
          & {\bm {\alpha_m}}= softmax(\bm {C_{m}^\top}\bm {u_m}), \\
          & \bm{\hat{m}}=\sigma(\bm {\hat{W}_m}\bm {C_{mn}}\bm {\alpha_m}+\bm {\hat{b}_m});\\
          \end{aligned}
          \right.
        \\
          &\left\{
          \begin{aligned}
          & \bm {C_n}=tanh(\bm{W_n}\bm{C_{mn}^\top}+\bm{b_n}),\\
          & {\bm {\alpha_n}}=softmax(\bm{C_{n}^\top}\bm{u_n}),\\
          & \bm{\hat{n}}=\sigma(\bm{\hat{W}_n}\bm{C_{mn}^\top}\bm {\alpha_n}+\bm{\hat{b}_n});\\
          \end{aligned}
          \right.
          \end{aligned}
        \\
        \bm u_{mn}=\bm {\hat m}\oplus \bm{\hat{n}};
      \end{gathered}
    \end{split}
  \end{equation}
\end{center}
% \begin{center}
% \begin{equation}
% \begin{split}
% \begin{gathered}
% \label{equ:cmf}
% C_{mn}=\bm m \times \bm n^\top ;\\
% \left\{
% \begin{aligned}
% C_i^m&=tanh(W_m^{\top}C_i^{mn\top}+b_m^{\top}),\\
% \alpha^m_i&= softmax(C_i^m), \\
% \end{aligned}
% \right.
% \quad i=0,1,..,d_m;\\
% \bm{\hat{m}'}=\sum_{d_m} \alpha_i^mC_i^{mnT},\quad 
% \bm{\hat{m}}=\sigma(\hat{W}_m^{\top}\bm{\hat{m}'}+\hat{b}_m^{\top});\\
% \left\{
% \begin{aligned}
% C_j^n&=tanh(W_n^{\top}C_j^{mn}+b_n^{\top}), \\
% \alpha_j^n&=softmax(C_j^n), \\
% \end{aligned}
% \right.
% \quad j=0,1,..,d_n;\\
% \bm{\hat{n}'}=\sum_{d_n} \alpha_j^nC_j^{mn},\quad
% \bm{\hat{n}}=\sigma(\hat{W}_n^{\top}\bm{\hat{n}'}+\hat{b}_n^{\top});\\
% \bm u_{mn}=\bm {\hat m}\oplus \bm{\hat{n}};
% \end{gathered}
% \end{split}
% \end{equation}
% \end{center}
where $\bm m, \bm n \in \{\bm t, \bm a, \bm v\}$ and $\bm m \ne \bm n$, $\sigma$ represents an activation function, 
% $d_m$ is the dimension of $\bm m$, $d_n$ is the dimension of $\bm n$, 
$\bm {\hat m}$ and $\bm {\hat n}$ refer to the reconstructed vectors of $\bm m $ and $ \bm n $, respectively, and the remaining variables stand for latent parameters learned from training. The final unified representation $\bm u$ is defined as $\bm u=\bm{u_{tv}}\oplus\bm{u_{ta}}\oplus\bm{u_{av}}$.
\subsection{Training and Optimization}
The Adam algorithm~\cite{kingmaadam} is used as optimizer to minimize the binary cross entropy function. The binary cross entropy is defined as follows:
\begin{equation}
\mathcal{L}=-\frac{1}{N}\sum_{n=1}^{N}[{y_n\log\hat{y}_n}+(1-y_n)\log(1-\hat{y}_n)],
\end{equation}
where $N$ denotes the number of target values, $y_n$ denotes the $n^{th}(1\le n \le N)$ true value, and $\hat{y}_n$ denotes the corresponding predicted value. For the information retrieval task, given a query, we randomly generate a fake candidate along with the correct candidate from the candidate corpus for effective training. 
\begin{comment}
{\color{blue}Training stops when model observes  that the loss keeps growing for 3 epochs or the iteration meets the max epoch. The batch-size is set as 64 empirically.}  
\end{comment}

\section{Experimental Evaluation}
Evaluation is conducted on two kinds of tasks, sentiment analysis and music information retrieval, as presented in Sections~\ref{sec:SA} and~\ref{sec:IR}, respectively\footnote{The source code and more details on the experimental settings are available at \url{http://github.com/mengshor/JTAV}}.

\subsection{Sentiment Analysis}\label{sec:SA}
%\subsubsection{Task Description}
{\textbf {Task Description}}. Sentiment analysis is a task that deals with the computational detection and extraction of opinions, beliefs, and emotions in given content~\cite{paltoglou2017sensing}. In this paper, we analyze posts from a social media platform ShutterSong. Given a post, which includes the visual (image), textual (song lyrics and user defined caption), and acoustic (song clip) parts, the goal is to tag a ``positive'' or ``negative'' label for the multi-modal content. We use ``1'' to indicate the positive label, and ``0'' to indicate the negative label. The sentiment analysis task can be regarded as a binary classification task.\par
\noindent {\textbf {Dataset and Parameter Settings}}. We sort a sub dataset from the ShutterSong dataset\footnote{\url{https://drive.google.com/file/d/0B2N8XiDRrEgISXFJSXBEMWpUMDA/view}.} to conduct the sentiment analysis task. The user-defined mood tags are used as labels, and 3260 items have available moods. After removing duplications, we obtain 272 mood tags. We divide these mood tags into ``positive'' and ``negative'' on the basis of the related meaning manually. We obtain 2297 positive and 963 negative samples, with each including a user posted image, a song clip, and its corresponding lyrics, part of the samples have available captions. We call this dataset as MoodShutter, and separate it into train (80\%), validation (10\%), and test (10\%) sets.\par
The song lyrics are truncated at 100 words, and captions are cleaned into five words according to a caption corpus. We use a 300 dimensional embedding matrix, which is pre-trained on the English Wikipedia dataset\footnote{\url{https://en.wikipedia.org/wiki/Wikipedia:Database_download#English-language_Wikipedia}.} through FastText, as ${W_e}$. The number of the hidden units of the BiGRU encoder is set as 50. Raw song clips are transformed into spectrograms with the \textit{librosa} tool~\cite{mcfee2017librosa}. We cut all the audio files into 10-second-segments. Audio sample rates are set as 22050Hz, hop lengths are 1024, and the numbers of frequency bins are 96. The parameter $N$ of DCRNN is set as $N=2$. The images are sized to $224\times 224\times 3$. \par
\noindent {\textbf {Baseline approaches}}. We compare our JTAV framework with:
\begin{itemize}
	\item doc2vec~\cite{le2014distributed}, a paragraph embedding approach; 
    %a paragraph embedding approach 
	\item lyricsHAN~\cite{alexandros2017lyrics}, a hierarchical attention network which encodes songs into vectors; 
	\item BiGRU, a bidirectional GRU network which utilizes the song lyrics as input; 
	\item attBiGRU, our novel text modeling approach which adds the available captions as the supplying role based on BiGRU; 
    \item MFCC~\cite{mcfee2017librosa}, a type of cepstral representation of the audio clip generated by \textit{librosa}; 
	\item convnet~\cite{choi2017transfer}, an acoustic feature learning method based on transfer learning; 
	\item DCRNN-MelS, our novel audio modeling approach which takes MelSs as input; 
	\item DCRNN-CQT, our novel audio modeling approach which takes CQTs as input; 
	\item F-VGG~\cite{simonyan2014very}, a VGG network pre-trained on the ImageNet dataset and fine tuned on the MoodShutter dataset; 
	\item F-ResNet~\cite{he2016deep}, a ResNet pre-trained on the ImageNet dataset and fine tuned on the MoodShutter dataset;
	\item F-DenseNet~\cite{huang2017densely}, a DenseNet pre-trained on the ImageNet dataset and fine tuned on the MoodShutter dataset; and	
	\item early fusion, a method which combines features learned by our methods before classification or regression tasks, and is very popularly used~\cite{chen2017visual,ramas2017multi}.
\end{itemize}
\begin{table}[t]
	\centering
	\caption{Results of the sentiment analysis task}
    \scalebox{0.8}{
	\begin{tabular}{lllccc}
		\toprule
		\multicolumn{1}{l}{modal} & \multicolumn{1}{l}{materials} & approaches & AUC score  & F1 score   & precision score\\
		\midrule
		\multicolumn{1}{l}{\multirow{5}{*}{text}} & \multicolumn{1}{l}{\multirow{4}{*}{lyrics}} & doc2vec~\cite{le2014distributed}  & {0.513} & {0.545} & {0.593 } \\
		\cmidrule{3-6}          &       & lyricsHAN~\cite{alexandros2017lyrics} &{0.518} &{0.575} & {0.596} \\
		\cmidrule{3-6}          &       & BiGRU & 0.572 & 0.602 & 0.640  \\
		\cmidrule{2-6}          & \multicolumn{1}{l}{{lyrics+caption}} & {attBiGRU} & {0.581} & {0.652} & {0.650 } \\
        \midrule
		\multicolumn{1}{l}{\multirow{5}{*}{audio}} & \multicolumn{1}{l}{\multirow{5}{*}{\mbox{song clip}}} & MFCC~\cite{mcfee2017librosa} & {0.505} & {0.549} & {0.586} \\
		\cmidrule{3-6}          &       & convnet~\cite{choi2017transfer} & {0.518} &{0.614} & {0.621 } \\
		\cmidrule{3-6}          &       & DCRNN-MelS& {0.536} & {0.635} & {0.634} \\
		\cmidrule{3-6}          &       & DCRNN-CQT & {0.559} &  {0.651} & {0.643} \\
		\midrule		
		\multicolumn{1}{l}{\multirow{4}{*}{image}} & \multicolumn{1}{l}{\multirow{4}{*}{image}} & F-VGG~\cite{simonyan2014very} & {0.546} & {0.594} & {0.619 } \\
		\cmidrule{3-6}          &       & F-ResNet~\cite{he2016deep} & {0.578} &{0.618} &{0.645 } \\
		\cmidrule{3-6}          &       & F-DenseNet~\cite{huang2017densely} & {0.588} & {0.627} & {0.653 } \\
		\midrule
		\multicolumn{1}{l}{\multirow{2}{*}{text+audio}} & \multicolumn{1}{l}{\multirow{2}{5em}{lyrics+caption+ song~clip}} & early fusion & \multirow{1}[2]{*}{0.586} & \multirow{1}[2]{*}{0.660} & \multirow{1}[2]{*}{0.656} \\
%		&       & \cite{chen2017visual,ramas2017multi}      &       &       &  \\
		\cmidrule{3-6}          &       & CMF-AP & 0.597 & 0.673 & 0.668 \\
		\midrule
		\multicolumn{1}{l}{\multirow{2}{*}{text+image}} & \multicolumn{1}{l}{\multirow{2}{5em}{lyrics+caption+ image}} & early fusion & \multirow{1}[2]{*}{0.593} & \multirow{1}[2]{*}{0.663} & \multirow{1}[2]{*}{0.661 } \\
%		&       & \cite{chen2017visual,ramas2017multi}    &       &       &  \\
		\cmidrule{3-6}          &       & CMF-AP & 0.611 & 0.688& 0.683 \\
		
		\midrule
		
		\multicolumn{1}{l}{\multirow{2}{*}{audio+image}} & \multicolumn{1}{l}{\multirow{2}{*}{image+\mbox{song clip}}} & early fusion & \multirow{1}[2]{*}{0.589} & \multirow{1}[2]{*}{{0.624}} & \multirow{1}[2]{*}{{0.653}} \\
%		&       &  \cite{chen2017visual,ramas2017multi}      &       &       &  \\
		\cmidrule{3-6}          &       & CMF-AP & 0.603 & {0.654} & {0.665}\\
		\midrule
		
		\multicolumn{1}{l}{\multirow{2}{*}{text+audio+image}} & \multicolumn{1}{l}{\multirow{2}{6em}{lyrics+caption+ image+\mbox{song clip}}} & early fusion & \multirow{1}[2]{*}{ 0.602 } & \multirow{1}[2]{*}{ 0.671 } & \multirow{1}[2]{*}{ 0.669 } \\
%		&       &  \cite{chen2017visual,ramas2017multi}     &       &       &  \\
		\cmidrule{3-6}          &       & CMF-AP & \textbf{ 0.623 }&  \textbf{0.691} &  \textbf{0.688} \\
		\midrule
		\multicolumn{6}{l}{{JTAV=attBiGRU+(F-DenseNet)+(DCRNN-CQT)+(CMF-AP)}} \\
		\bottomrule
	\end{tabular}%
    }
	\label{tab:results}%
\end{table}%
\noindent {\textbf {Evaluation Metrics}}. The positive and negative samples are unbalanced and the ratio is about $7:3$. The weighted average area under the curve~(AUC) score, F1 score, and precision score are chosen as evaluation metrics\footnote{We utilize the implementation in sklearn, \url{http://scikit-learn.org/}.}.\par
\noindent {\textbf {Results and Discussion}}. Table~\ref{tab:results} presents the results between JTAV and the baseline approaches. For fair comparison, all feature vectors are obtained through optimized training; the classic logistic regression approach is utilized as the classier; and the reported results are the average values of 10 runs.The values in boldface represent the best results among all the approaches. The proposed JTAV performed the best on all metrics, with 0.623 AUC score, 0.691 F1 score, 68.8\% precision score. \par
\noindent \textit{Observations on single modality}. For textual content, attBiGRU obtained the best results, and was superior to doc2vec with about 7 percentage points promotion of AUC score, 11 percentage points promotion of F1 score, and 6 percentage points promotion of precision score. These results demonstrate the proposed BiGRU approach extracts more powerful features than doc2vec and lyricsHAN. And taking the omni textual component into consideration, the attBiGRU approach can generate better results. For acoustic content, DCRNN outperforms MFCC and convnet because it can learn acoustic information both locally and globally. Long-term temporal context dependency is also retained.We also observe that CQT works better than MelS. DenseNet is the best choice for encoding images into vectors as compared with VGG and ResNet. \par
\noindent \textit{Observations on multiple modalities}. When more modalities emerge, more information is included, and better results are obtained. For example, the performance of combining textual and acoustic information is much better than that of using texts or audio solely. Our CMF-AP approach works steadily better than the baseline approach, because it not only extracts the inner features inside single modality, but also the the outer information cross modalities. Moreover, the best results are presented when we utilize all modalities and all available materials~(i.e., JTAV). The AUC score is 0.623, representing an improvement of 2.1 percentage points to the early fusion approach. The F1 score is 0.691, performing an improvement of 2 percentage points over the best baseline approach. The precision score is 0.688, representing about 1.9 percentage points improvement to the best baseline approach. It seems like that the promotions of the proposed JTAV over the best baseline approach is not that huge. As a matter of fact, the early fusion approach utilizes the effective features generated by our proposed work~(i.e., attBiGRU, DCRNN-CQT, and F-DenseNet). We believe that, if compared to the original version of early fusion method which only uses the previous features, the margin would be larger. For the sake of space limitation, in Table 1 the early fusion approach on previous features are omitted.

\subsection{Music Information Retrieval}\label{sec:IR}

We conduct an extended experiment, which is can be explained as given an image query and named as image2song~\cite{li2017image2song}, to verify the general effectiveness of JTAV. Image2song is a music information retrieval task and aims to find the most relevant song.\par
% †§
We perform the image2song task on two benchmark datasets, they are the Shuttersong† dataset and the Shuttersong§ dataset~\cite{li2017image2song}. Both datasets include no-repeat 620 songs with 3100 images, that is, each song is related to five images. Part of the images are attached with user-defined captions, and each song has an acoustic song clip and corresponding song lyrics. The difference lies in the partition strategy of the train and test dataset. In Shuttersong†, 100 songs and related images are selected randomly for testing, and the rest for training. While in Shuttersong§, the train and test set share the whole song set, while one of the five images of each song is chosen randomly for testing. The preprocessing settings of images, captions, song lyrics and song clips is the same as that of the sentiment analysis task.\par
We utilize the rank-based evaluation metrics to compare the results. For fair comparison, we adopt the evaluate metrics in~\cite{li2017image2song}. Med r represents the medium rank of the ground truth retrieved song, and lower values indicate better performance. Recall@k~(R@k for short), is the percentage of a ground truth song retrieved in the top-k ranked items, and higher values indicate better performance. In dataset†, $k=\{1,5,10\}$, and $1\le$ Med r $\le100$; whereas in Shuttersong§,  $k=\{10,50,100\}$, and $1\le$ Med r $\le620$.\par
\begin{figure}[t]
	\centering
	\begin{minipage}[t]{0.48\textwidth}
		\centering
		\includegraphics[width=7.6cm]{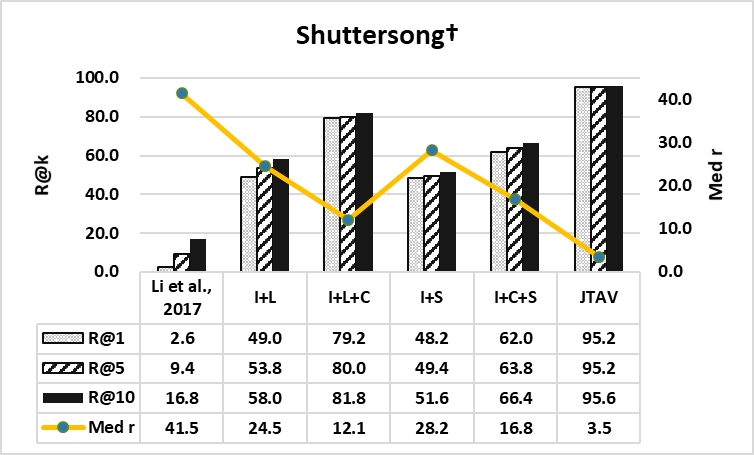}
		\caption{Results of image2song on Shuttersong†}
		\label{fig:dataset_1}
	\end{minipage}
	\begin{minipage}[t]{0.48\textwidth}
		\centering
		\includegraphics[width=7.6cm]{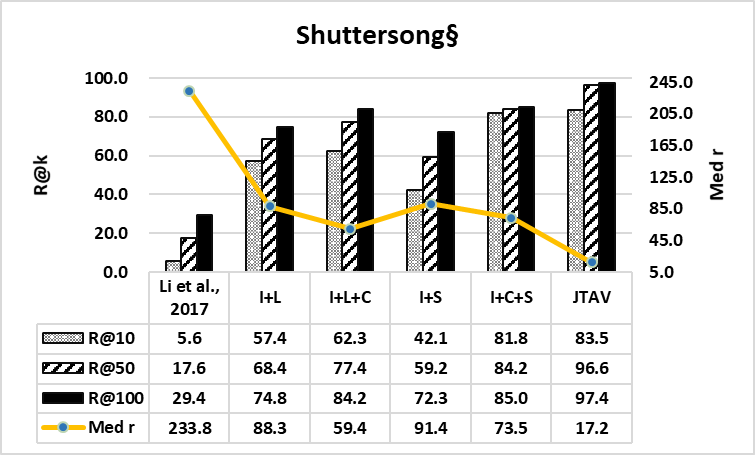}
		\caption{Results of image2song on Shuttersong§}
		\label{fig:dataset_2}    
	\end{minipage}	
\end{figure}
%Since the captions has no more than 5 words, 
We compare the proposed JTAV with the state-of-the-art approach~\cite{li2017image2song}. For a more comprehensive comparison, we design several baseline approaches based on our proposed strategies: I+L uses images as queries and song lyrics as candidates with a suit of BiGRU, F-DenseNet and CMF-AP; I+C+L uses images and available captions as queries while song lyrics as candidates with a suit of attBiGRU, F-DenseNet and CMF-AP; I+S uses images as queries and song clips as candidates with a suit of DCRNN-CQT, F-DenseNet and CMF-AP; and I+C+S uses images and available captions as queries while song clips as candidates with a suit of DCRNN-CQT, F-DenseNet and CMF-AP. In JTAV, the images and available captions form the queries, the song clips and lyrics form the candidates.\par

Fig.~\ref{fig:dataset_1} and Fig.~\ref{fig:dataset_2} display the results on Shuttersong† and Shuttersong§, respectively. Both prove JTAV is the most powerful one among all testing approaches. JTAV turns the image2song task from finding the most similar item to finding the most relevant item, which is nearer the real situation in cross-modal retrieval tasks. Notably, JTAV exceeds the state-of-the-art approaches soundly. In Shuttersong†, the Med r is no more than 5, and the R@1 score is more than 95\%, which is more than 90\% better than that obatined in~\cite{li2017image2song}. In Shuttersong§,  the Med r is no more than 20, which is less a tenth of that obtained in~\cite{li2017image2song}, the R@10 score is more than 83\%, and the  R@100 score is more than 97\%.

\section{Conclusion}
In this paper, we aim to address the issue of learning social media content from the multi-modal view. We have designed effective approaches~(i.e., attBiGRU) for textual content, F-DenseNet for visual content, and DCRNN for acoustic content, to extract fine-grained features. We have introduced CMF-AP to generate cross-modal aware matrices and reconstructed modal vectors, which are used to produce an unified representation of the multi-modal content. The proposed framework has been intensively evaluated, and the experimental results have demonstrated the general validity of JTAV by comparing with the state-of-the-art approaches.

\section*{Acknowledgements}
This work was supported in part by the National Natural Science Foundation of China under Grant No.U1636116, 11431006, 61772288, the Research Fund for International Young Scientists under Grant No. 61650110510 and 61750110530, and the Ministry of education of Humanities and Social Science project under grant 16YJC790123.

%% file: jvta-coling.bbl
\begin{thebibliography}{}

\bibitem[\protect\citename{Agichtein \bgroup et al.\egroup
  }2008]{agichtein2008finding}
Eugene Agichtein, Carlos Castillo, Debora Donato, Aristides Gionis, and Gilad
  Mishne.
\newblock 2008.
\newblock Finding high-quality content in social media.
\newblock In {\em Proceedings of the international conference on web search and
  data mining}, pages 183--194. ACM.

\bibitem[\protect\citename{Alexandros}2017]{alexandros2017lyrics}
Tsaptsinos Alexandros.
\newblock 2017.
\newblock Lyrics-based music genre classification using a hierarchical
  attention network.
\newblock In {\em Proceedings of the 18th International Society of Music
  Information Retrieval Conference}, pages 694--701. ISMIR.

\bibitem[\protect\citename{Boashash}1996]{boashash1996time}
Boualem Boashash.
\newblock 1996.
\newblock Time frequency signal analysis: {Past}, present and future trends.
\newblock In {\em Control and Dynamic Systems}, volume~78, pages 1--69.
  Elsevier.

\bibitem[\protect\citename{Bojanowski \bgroup et al.\egroup
  }2017]{bojanowski2016enriching}
Piotr Bojanowski, Edouard Grave, Armand Joulin, and Tomas Mikolov.
\newblock 2017.
\newblock Enriching word vectors with subword information.
\newblock {\em Transactions of the Association for Computational Linguistics},
  5:135--146.

\bibitem[\protect\citename{Brown and Puckette}1992]{brown1992efficient}
Judith~C Brown and Miller~S Puckette.
\newblock 1992.
\newblock An efficient algorithm for the calculation of a constant {Q}
  transform.
\newblock {\em The Journal of the Acoustical Society of America},
  92(5):2698--2701.

\bibitem[\protect\citename{Chen \bgroup et al.\egroup }2017]{chen2017visual}
Xingyue Chen, Yunhong Wang, and Qingjie Liu.
\newblock 2017.
\newblock Visual and textual sentiment analysis using deep fusion convolutional
  neural networks.
\newblock In {\em Proceedings of the 2017 IEEE International Conference on
  Image Processing}, pages 296--300. IEEE.

\bibitem[\protect\citename{Choi \bgroup et al.\egroup }2017a]{choi2017transfer}
Keunwoo Choi, George Fazekas, Mark Sandler, and Kyunghyun Cho.
\newblock 2017a.
\newblock Transfer learning for music classification and regression tasks.
\newblock In {\em Proceedings of the International Society of Music Information
  Retrieval Conference}. ISMIR.

\bibitem[\protect\citename{Choi \bgroup et al.\egroup }2017b]{choi2017tutorial}
Keunwoo Choi, Gy{\"{o}}rgy Fazekas, Kyunghyun Cho, and Mark~B. Sandler.
\newblock 2017b.
\newblock A tutorial on deep learning for music information retrieval.
\newblock {\em CoRR}, abs/1709.04396.

\bibitem[\protect\citename{Garimella \bgroup et al.\egroup
  }2016]{garimella2016social}
Venkata Rama~Kiran Garimella, Abdulrahman Alfayad, and Ingmar Weber.
\newblock 2016.
\newblock Social media image analysis for public health.
\newblock In {\em Proceedings of the CHI Conference on Human Factors in
  Computing Systems}, pages 5543--5547. ACM.

\bibitem[\protect\citename{Han \bgroup et al.\egroup }2013]{han2013lexical}
Bo~Han, Paul Cook, and Timothy Baldwin.
\newblock 2013.
\newblock Lexical normalization for social media text.
\newblock {\em ACM Transactions on Intelligent Systems and Technology}, 4(1):5.

\bibitem[\protect\citename{Huang \bgroup et al.\egroup }2017]{huang2017densely}
Gao Huang, Zhuang Liu, Laurens van~der Maaten, and Kilian~Q Weinberger.
\newblock 2017.
\newblock Densely connected convolutional networks.
\newblock In {\em Proceedings of the IEEE Conference on Computer Vision and
  Pattern Recognition}, pages 4700--4708.

\bibitem[\protect\citename{Hutto and Gilbert}2014]{hutto2014vader}
Clayton~J Hutto and Eric Gilbert.
\newblock 2014.
\newblock Vader: A parsimonious rule-based model for sentiment analysis of
  social media text.
\newblock In {\em Proceedings of the 8th international AAAI conference on
  weblogs and social media}.

\bibitem[\protect\citename{Jansen \bgroup et al.\egroup }2017]{aren2017towards}
Aren Jansen, Manoj Plakal, Ratheet Pandya, Dan Ellis, Shawn Hershey, Jiayang
  Liu, Channing Moore, and Rif~A. Saurous.
\newblock 2017.
\newblock Towards learning semantic audio representations from unlabeled data.
\newblock In {\em Proceedings of Workshop on Machine Learning for Audio Signal
  Processing at the 31st Conference on Neural Information Processing Systems}.
  NIPS.

\bibitem[\protect\citename{Kingma and Ba}2015]{kingmaadam}
Diederik~P. Kingma and Jimmy~Lei Ba.
\newblock 2015.
\newblock Adam: A method for stochastic optimization.
\newblock In {\em Proceedings of the 3th International Conference on Learning
  Representations}. ICLR.

\bibitem[\protect\citename{Le and Mikolov}2014]{le2014distributed}
Quoc Le and Tomas Mikolov.
\newblock 2014.
\newblock Distributed representations of sentences and documents.
\newblock In {\em Proceedings of International Conference on Machine Learning},
  pages 1188--1196.

\bibitem[\protect\citename{Lei \bgroup et al.\egroup }2017]{lei2017swim}
Wenqiang Lei, Xuancong Wang, Meichun Liu, Ilija Ilievski, Xiangnan He, and
  Min-Yen Kan.
\newblock 2017.
\newblock Swim: a simple word interaction model for implicit discourse relation
  recognition.
\newblock In {\em Proceedings of the 26th International Joint Conference on
  Artificial Intelligence}, pages 4026--4032.

\bibitem[\protect\citename{Lei \bgroup et al.\egroup }2018a]{lei2018sequicity}
Wenqiang Lei, Xisen Jin, Min-Yen Kan, Zhaochun Ren, Xiangnan He, and Dawei Yin.
\newblock 2018a.
\newblock Sequicity: Simplifying task-oriented dialogue systems with single
  sequence-to-sequence architectures.
\newblock In {\em Proceedings of the 56th Annual Meeting of the Association for
  Computational Linguistics (Volume 1: Long Papers)}, pages 1437--1447.

\bibitem[\protect\citename{Lei \bgroup et al.\egroup }2018b]{lei2018linguistic}
Wenqiang Lei, Yuanxin Xiang, Yuwei Wang, Qian Zhong, Meichun Liu, and Min-Yen
  Kan.
\newblock 2018b.
\newblock Linguistic properties matter for implicit discourse relation
  recognition: Combining semantic interaction, topic continuity and
  attribution.
\newblock In {\em Proceedings of the AAAI Conference on Artificial
  Intelligence}, volume~32.

\bibitem[\protect\citename{Li \bgroup et al.\egroup }2017]{li2017image2song}
Xuelong Li, Di~Hu, and Xiaoqiang Lu.
\newblock 2017.
\newblock Image2song: Song retrieval via bridging image content and lyric
  words.
\newblock In {\em Proceedings of IEEE International Conference on Computer
  Vision}, pages 5650--5659.

\bibitem[\protect\citename{McFee \bgroup et al.\egroup }2017]{mcfee2017librosa}
Brian McFee, Matt McVicar, Oriol Nieto, Stefan Balke, Carl Thome, Dawen Liang,
  Eric Battenberg, Josh Moore, Rachel Bittner, Ryuichi Yamamoto, et~al.
\newblock 2017.
\newblock librosa 0.5.0.

\bibitem[\protect\citename{Oramas \bgroup et al.\egroup }2017]{ramas2017multi}
Sergio Oramas, Oriol Nieto, Francesco Barbieri, and Xavier Serra.
\newblock 2017.
\newblock Multi-label music genre classification from audio, text, and images
  using deep features.
\newblock In {\em Proceedings of the 18th International Society of Music
  Information Retrieval Conference}. ISMIR.

\bibitem[\protect\citename{Paltoglou and Thelwall}2017]{paltoglou2017sensing}
Georgios Paltoglou and Mike Thelwall.
\newblock 2017.
\newblock Sensing social media: a range of approaches for sentiment analysis.
\newblock In {\em Cyberemotions}, pages 97--117. Springer.

\bibitem[\protect\citename{Panwar \bgroup et al.\egroup }2017]{panwar2017deep}
Sharaj Panwar, Arun Das, Mehdi Roopaei, and Paul Rad.
\newblock 2017.
\newblock A deep learning approach for mapping music genres.
\newblock In {\em Proceedings of the 12th System of Systems Engineering
  Conference}, pages 1--5. IEEE.

\bibitem[\protect\citename{Park and Im}2016]{park2016image}
Gwangbeen Park and Woobin Im.
\newblock 2016.
\newblock Image-text multi-modal representation learning by adversarial
  backpropagation.
\newblock {\em CoRR}, abs/1612.08354.

\bibitem[\protect\citename{Simonyan and Zisserman}2015]{simonyan2014very}
Karen Simonyan and Andrew Zisserman.
\newblock 2015.
\newblock Very deep convolutional networks for large-scale image recognition.
\newblock In {\em Proceedings of the 3th International Conference on Learning
  Representations}. ICLR.

\bibitem[\protect\citename{Wang \bgroup et al.\egroup }2013]{wang2013social}
Zhi Wang, Wenwu Zhu, Peng Cui, Lifeng Sun, and Shiqiang Yang.
\newblock 2013.
\newblock Social media recommendation.
\newblock In {\em Social Media Retrieval}, pages 23--42. Springer.

\bibitem[\protect\citename{Wang \bgroup et al.\egroup
  }2015]{wang2015unsupervised}
Yilin Wang, Suhang Wang, Jiliang Tang, Huan Liu, and Baoxin Li.
\newblock 2015.
\newblock Unsupervised sentiment analysis for social media images.
\newblock In {\em Proceedings of the International Conference on Artificial
  Intelligence}, pages 2378--2379. AAAI Press.

\bibitem[\protect\citename{Wu \bgroup et al.\egroup }2017]{wu2017mobile}
Chao Wu, Yaoxue Zhang, Jia Jia, and Wenwu Zhu.
\newblock 2017.
\newblock Mobile contextual recommender system for online social media.
\newblock {\em IEEE Transactions on Mobile Computing}, 16(12):3403--3416.

\bibitem[\protect\citename{Yang \bgroup et al.\egroup
  }2016]{yang2016hierarchical}
Zichao Yang, Diyi Yang, Chris Dyer, Xiaodong He, Alexander~J Smola, and
  Eduard~H Hovy.
\newblock 2016.
\newblock Hierarchical attention networks for document classification.
\newblock In {\em Proceedings of the 14th Annual Conference of the North
  American Chapter of the Association for Computational Linguistics: Human
  Language Technologies}, pages 1480--1489. NAACL.

\bibitem[\protect\citename{Zhu \bgroup et al.\egroup }2017]{zhu2017fusing}
Bilei Zhu, Fuzhang Wu, Ke~Li, Yongjian Wu, Feiyue Huang, and Yunsheng Wu.
\newblock 2017.
\newblock Fusing transcription results from polyphonic and monophonic audio for
  singing melody transcription in polyphonic music.
\newblock In {\em Proceedings of the 2017 IEEE International Conference on
  Acoustics, Speech and Signal Processing}. IEEE.

\end{thebibliography}
